\documentclass[11pt]{article}

\usepackage{acl}

\usepackage{times}
\usepackage{latexsym}
\usepackage{graphicx}
\usepackage{amsmath}
\usepackage{amssymb}
\usepackage{booktabs}
\usepackage{multirow}
\usepackage{makecell}
\usepackage{xcolor}
\usepackage{microtype}
\usepackage{xspace}

\newcommand{\method}{MoE$^2$-LoRA\xspace}

\title{MoE$^2$-LoRA: When MoE Models Meet MoE-style Low-Rank Adaptation}

\author{
{\normalfont Qingyu Yang$^{1,2,*}$ \quad
Haonan He$^{1,3,*,\dagger}$ \quad
Minglei Li$^{1,4}$ \quad
Jingqi Ye$^{1,3}$} \\
{\normalfont Tao Chen$^{1}$ \quad
Lei Bai$^{1}$ \quad
Peng Ye$^{1,4,5,\ddagger}$} \\
{\normalfont\small $^{1}$Shanghai Artificial Intelligence Laboratory \quad
$^{2}$KTH Royal Institute of Technology} \\
{\normalfont\small $^{3}$University of Science and Technology of China \quad
$^{4}$Fudan University} \\
{\normalfont\small $^{5}$The Chinese University of Hong Kong} \\
{\normalfont\small * Equal contribution. \quad
$\dagger$ Project Lead. \quad
$\ddagger$ Corresponding author.} \\
{\normalfont\small Email: \texttt{yepeng@pjlab.org.cn}}
}

\begin{document}
\maketitle

\begin{abstract}
Mixture-of-Experts (MoE) architectures have been widely adopted in large language models, yet parameter-efficient fine-tuning (PEFT) for MoE models remains underexplored. 
Existing PEFT methods for MoE either ignore router priors with uniform adapters, reducing efficiency and risking forgetting, or rely on static expert selection, limiting per-token capacity and cross-expert feature learning.
In this paper, we make the first attempt to fine-tune MoE models with MoE-style low-rank adaptation: 
our method, entitled \method, deeply couples the pretrained expert specialization with task-specific adaptivity via a dual-channel Routing-Conditioned Projection (RCP) module, which reuses base router activations to inform LoRA routing. We further introduce a single global LoRA expert pool shared across all layers, 
enabling model-wide adaptation with emergent layer-wise affinities and balanced expert utilization.
\method simultaneously benefits from the advantages of prior reuse, dynamic adapter routing, and model-wide knowledge sharing.
Evaluated on multiple MoE backbones with varying scales and expert granularities, \method consistently achieves state-of-the-art downstream accuracy while retaining stronger general capabilities.
\end{abstract}

\section{Introduction}
Mixture-of-Experts (MoE)~\citep{jacobs1991adaptive} architectures, which activate only a small fraction of parameters for each input, have recently been widely adopted by 
large language models (LLMs) such as DeepSeek-V3~\citep{liu2024deepseek} and Qwen3~\cite{yang2025qwen3}, owing to their advantages including reasoning efficiency and scaling performance comparable to that of dense models. For dense architecture based LLMs, efficient and effective fine-tuning methods have been thoroughly studied. For example, LoRA~\citep{hu2021lora}, one of the most widely used parameter-efficient fine-tuning (PEFT) methods designed for dense models, matches the performance of full fine-tuning by decomposing weight updates into low-rank subspaces. In comparison, how to efficiently fine-tune MoE models remains an underexplored problem, primarily due to their sparse activation nature, unstable gradient flow, and massive parameter scale, which highlights the pressing need for PEFT methods specifically tailored for MoE models.

\begin{figure*}[!t]
\centering
\includegraphics[width=\textwidth]{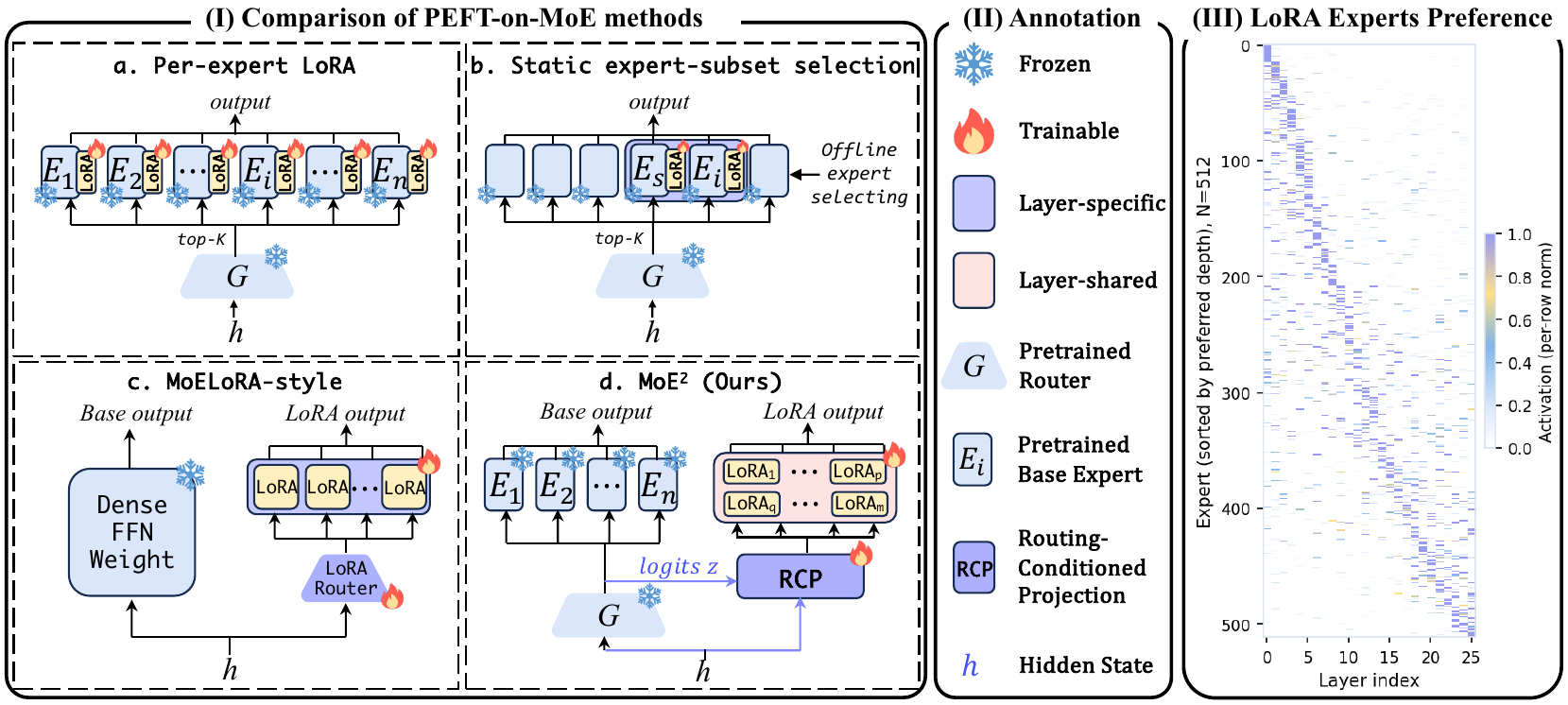}
\caption{(I) Comparison of PEFT-on-MoE methods: (a) Per-expert LoRA, which equip each MoE expert with a LoRA adapter; (b) Static expert-subset selection methods, which select a subset of experts offline to be trained with LoRA; (c) MoELoRA-style methods, which adapt each MoE module with a MoE-style low-rank adapter; (d) MoE$^2$. (II) Annotation. (III) LoRA Experts Preference: We present the preferred layers of LoRA experts in MoE$^2$'s global expert pool, demonstrating the layer-specific learning and cross-layer learning capabilities of MoE$^2$. }
\label{fig:overview}
\vspace{-3mm}
\end{figure*}

Existing PEFT methods for MoE models are predominantly built upon applying LoRA adapters to MoE experts (Fig.~\ref{fig:overview}(I.a)), and further attempt to exploit the routing priors of the MoE model to guide expert-level adaptation, that is, to decide which experts should receive adaptation or how much (Fig.~\ref{fig:overview}(I.b)).
For example, ESFT~\citep{esft2024} and DAS-LoRA~\citep{tang2026exploring} first collect offline router statistics to identify a subset of critical experts, and then apply LoRA only to those experts; DR-LoRA~\citep{deng2026dr} adaptively assigns higher ranks to experts with higher routing frequency and gradient-based importance. 
While these methods improve parameter efficiency, 
they still suffer from two major limitations: 
(i) Insufficient adaptation of critical experts. By selecting a subset of trainable experts or high-rank experts, any expert who falls outside this subset but is activated at inference time receives insufficient task-specific updates.
(ii) Lack of high-level interactions: Adapting experts independently neglects both the cooperative dynamics among co-activated experts
and the cross-layer 
interactions of MoE models 
(Fig.~\ref{fig:cka}), which could enhance performance and reduce parameter redundancy.

To address these limitations, we seek to fine-tune MoE models with dedicated MoE mechanisms, shifting the adaptation horizon from individual experts to each MoE module and the entire MoE model.
This direction resonates with another line of research that integrates LoRA with an MoE structure (Fig.~\ref{fig:overview}(I.c)), which yields MoE-style LoRA methods such as MoELoRA~\citep{luo2024moelora} and MoLA~\citep{gao2025mola}. These methods, however, are not specifically designed for MoE models: their MoE mechanisms operate independently of those inherent to MoE models, thereby leaving the expert interactions and cross-layer interactions encoded in the base MoE models unutilized.
In light of this, if it is possible to deeply bind the two MoE mechanisms of MoE models and MoE-style LoRA, 
we could mitigate the aforementioned limitations. 
These observations naturally raise a unifying question:
\textit{\textbf{When MoE models meet MoE-style low-rank adaptation, how can we deeply bind the two MoE mechanisms at both the MoE module level and the MoE model level, thereby enhancing task performance while preserving general capabilities?}}

To answer this question, we introduce \method (Figure~\ref{fig:overview}(I.d)), an MoE-style low-rank adaptation method tailored for MoE models. At its core lies the Routing-Conditioned Projection (RCP) module, which reuses the pretrained MoE routing activations to drive LoRA expert selection. Concretely, RCP fuses two routing signals: a primary channel \(z\) carrying the base router activations, along with an auxiliary channel \(h_c\), which injects a task-specific correction. A learnable matrix \(W_l\) maps the concatenated input into the LoRA routing space, producing LoRA expert selection coupled with the base model's expert selection. Further, in order to enable model-level knowledge sharing, we maintain a \textbf{global pool} of LoRA experts shared across all MoE layers. As demonstrated in Figure~\ref{fig:overview}(III), this design yields a striking emergent property: each LoRA expert develops a pronounced layer affinity, while still contributing to other layers to a lesser extent, and the overall assignment of experts across layers remains highly uniform. This indicates that the global pool achieves model-wide cross-layer knowledge sharing with high parameter utilization, avoiding the isolation and redundancy of layer-specific adapters. 

As a result, our \method unifies three key properties: pretrained routing prior reuse, MoE-style LoRA expert selection, and model-level cross-layer learning through a global expert pool. Equipped with this model-level design, \method attains state-of-the-art downstream accuracy on four MoE backbones of varying scale and expert granularity, OLMoE-1B-7B~\cite{olmoe2024}, DeepSeek-V2-Lite~\cite{liu2024deepseek}, Qwen3-30B-A3B~\cite{yang2025qwen3}, and Qwen3.5-35B-A3B~\cite{qwen3.5}, improving over the strongest PEFT baselines by up to +2.56 points in in-domain average accuracy, while exhibiting superior general capability retention. Our contributions are summarized as follows:

\begin{itemize}
\item We diagnose two key limitations of existing PEFT methods for MoE models and introduce the principle of fine-tuning MoE models with MoE-style PEFT, which requires deeply binding the two MoE mechanisms and establishing both module level and model level interactions. This principle unifies expert-level, module-level, and model-level adaptation, overcoming the identified limitations and serving as the foundation for our method design.

\vspace{-1.5mm}

\item  We propose \method, which realizes this insight through two key components: (1) a Routing-Conditioned Projection (RCP) module that repurposes the base router's activations to drive LoRA expert selection, and (2) a single global LoRA expert pool shared across all MoE layers. As visualized in Figure~\ref{fig:overview}(I), the global pool yields emergent layer affinities and balanced expert utilization, achieving efficient model-wide knowledge sharing.
\vspace{-1.5mm}

\item Across four MoE backbones (OLMoE-1B-7B, DeepSeek-V2-Lite, Qwen3-30B-A3B, Qwen3.5-35B-A3B) and various benchmarks (math, code, general, multimodal), \method achieves state-of-the-art downstream performance and general capability retention, improving over the base model in both task and general performance.
\end{itemize}

\section{Related Work}

\subsection{PEFT and LoRA.}
The scaling of LLMs to billions of parameters has rendered conventional full fine-tuning computationally prohibitive. This bottleneck has spurred the widespread adoption of PEFT techniques, epitomized by LoRA, owing to its plug-and-play deployability, strong downstream performance, and zero inference latency. By restricting updates to a small subset of trainable parameters, these methods enable resource-efficient model adaptation while preserving the model's original inference cost. Though LoRA matches the performance of full fine-tuning in many scenarios, a gap persists on challenging tasks such as scientific reasoning and coding. This gap has been extensively investigated and effectively narrowed by a recent body of follow-up work that refines LoRA with improved algorithms. For example, LoRA+~\citep{hayou2024lora+} decouples the learning rates for the down- and up-projection weights of a low-rank adapter, effectively enhancing training stability over standard LoRA. GoRA~\citep{he2026gora} adaptively assigns ranks to adapters based on gradient information, thereby strengthening parameter utilization. Similarly, methods such as MoELoRA~\citep{luo2024moelora}, MoLA~\citep{gao2025mola}, and LoRAMoE~\citep{dou2023loramoe} improve LoRA's performance by integrating MoE mechanisms. 

\subsection{PEFT for MoE models.}
PEFT methods are predominantly implemented and evaluated on dense models, without being tailored to MoE architectures. Existing studies on PEFT methods for MoE models remain sparse and insufficiently comparative. Though PERFT~\cite{liu2024perft} has briefly investigated various strategies for training MoE models with LoRA, most PEFT methods for MoE models concentrate on selecting experts to be fine-tuned or adaptively assigning ranks to the adapters of MoE experts based on routing statistics and expert importance information. For example, ESFT~\citep{esft2024} analyzes the router's output probability distribution on downstream task data and accordingly selects a subset of experts to be fine-tuned; MoE-Sieve~\cite{manzoni2026moe} counts how many tokens each expert receives per layer on a small calibration dataset, then applies LoRA adapters exclusively to the selected experts; CEFT~\citep{bai2025understanding} identifies experts that progressively amplify attention to relevant contextual information and selectively fine-tunes only these experts; DR-LoRA~\cite{deng2026dr} assigns heterogeneous LoRA ranks across experts based on an expert saliency score that jointly considers routing frequency and gradient-based rank importance.

\section{Method}
\label{sec:method}

\subsection{MoE Routing and LoRA}

An MoE layer contains a router $G(\cdot)$ and $N_E$ feed-forward
experts $\{E_i\}_{i=1}^{N_E}$. Given a token representation
$h\in\mathbb{R}^D$, the router produces logits $z=G(h)$ and activates
a top-$K$ expert set $\mathcal{S}_h$. The MoE  output is
\[
\mathrm{MoE}(h)=\sum_{i\in\mathcal{S}_h} g_i(h)E_i(h),
\]
where $g_i(h)$ is the normalized routing weight.

LoRA parameterizes a low-rank update to a pretrained weight
$W\in\mathbb{R}^{d_\mathrm{out}\times d_\mathrm{in}}$ as
\[
\Delta W=\frac{\alpha}{r}BA,
\]
where $A\in\mathbb{R}^{r\times d_\mathrm{in}}$,
$B\in\mathbb{R}^{d_\mathrm{out}\times r}$, and
$r\ll\min(d_\mathrm{out},d_\mathrm{in})$. In PEFT-on-MoE, the
pretrained router and experts are frozen, and only the introduced
adapter parameters are optimized.

\subsection{Routing-Conditioned Projection (\texorpdfstring{$W_l$}{Wl})}
\label{sec:wl}

\begin{figure}[t]
\centering
\includegraphics[width=0.98\columnwidth]{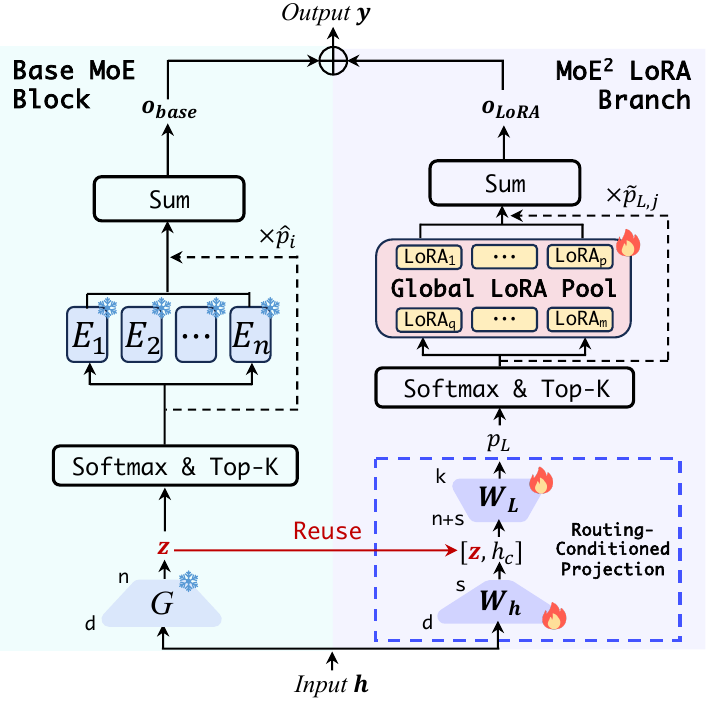}
\caption{
Overview of the \method forward pass on a single
MoE layer.}
\label{fig:method}
\vspace{-3mm}
\end{figure}

\method introduces an additional LoRA expert pool $\{E_j^{\mathrm{LoRA}}\}_{j=1}^{N_L}$ and learns per-token routing $p_L \in \mathbb{R}^{N_L}$ over this pool. The key design choice is the \emph{routing input}: rather than learning routing from raw hidden state $h$ or hardwiring it to base routing, \method uses a \emph{two-channel input}: (1) Primary channel: the base router logits $z = G(h) \in \mathbb{R}^{N_E}$. This provides a directly pretrained routing signal for LoRA expert selection instead of learning routing entirely from scratch; (2) Auxiliary channel: a low-rank projection of the hidden state, $h_c = W_h \cdot h \in \mathbb{R}^{d_b}$ with $d_b \ll D$. This adds a learn from scratch signal that complements the base router channel. The LoRA routing is computed as:
\begin{equation}
p_L \;=\; \mathrm{softmax}\bigl(W_l \cdot [\, z \,;\, h_c\,]\bigr),
\label{eq:p_L}
\end{equation}
where $W_l \in \mathbb{R}^{N_L \times (N_E + d_b)}$ is a learnable per-layer projection matrix. We then select the top-$K_L$ LoRA experts, $\mathcal{S}_h^{\mathrm{LoRA}} = \mathrm{TopK}(p_L)$, and renormalize their weights to sum to one: $\tilde{p}_{L,j} = p_{L,j} / \sum_{j' \in \mathcal{S}_h^{\mathrm{LoRA}}} p_{L,j'}$. Fig.~\ref{fig:method} illustrates the complete forward pass.

This design conditions LoRA routing on pretrained routing signals while preserving the flexibility of task-specific routing adaptation through the learnable projection $W_l$.

\subsection{Global Shared LoRA Pool}
\label{sec:global-pool}

\method's LoRA expert pool is \emph{shared across layers}: a set of $N_L$ LoRA experts is reused by multiple MoE blocks, with each block accessing the pool through its own per-layer $W_l$.

This contrasts with heuristic allocation-based designs such as MoLA, which modifies layer heterogeneity by \emph{statically pre-specifying} each layer's expert pool size and composition, an architectural choice that requires search over per-layer capacity schemes. \method instead handles layer heterogeneity \emph{dynamically at the routing level}: each layer can flexibly select from the shared pool through its learnable $W_l$, with no per-layer capacity hyperparameter. The same expert can serve multiple layers, and the same layer can route differently from its neighbors.

An evident advantage of cross-layer expert sharing is that the parameter count of the trainable LoRA experts pool do not scale linearly with model depth, in contrast to per-layer pool designs where the parameter count grows as $L \times N_{\mathrm{per\text{-}layer}}$. The choice of the total number of LoRA experts and the number of activated experts per input and per layer is clearly more flexible than that of the per-layer pool designs.

\subsection{Forward Pass}

Combining the RCP routing of \S\ref{sec:wl} with the global shared pool of \S\ref{sec:global-pool}, the \method block output is
\begin{equation}
y \;=\; \mathrm{MoE}(h) \;+\; \sum_{j \in \mathcal{S}_h^{\mathrm{LoRA}}} \tilde{p}_{L,j} \cdot E_j^{\mathrm{LoRA}}(h),
\label{eq:rcp_output}
\end{equation}
where $E_j^{\mathrm{LoRA}}(h) = (\alpha/r)\,B_j A_j h$ is the rank-$r$ LoRA expert, and $\tilde{p}_{L,j}$ are the renormalized top-$K_L$ routing weights defined in \S\ref{sec:wl} (or its hierarchical variant when $G > 1$).

\section{Experiments}

In this section, we evaluate \method and baseline methods on multiple MoE backbones across various tasks (math, code, general-retention, and multimodal).

\label{sec:experiments}
\subsection{Main Experiments}

\paragraph{Base Models.}
We conduct the main experiments on three MoE backbones: \textbf{OLMoE-1B-7B}~\citep{olmoe2024}, \textbf{DeepSeek-V2-Lite}~\citep{deepseekv2}, and \textbf{Qwen3-30B-A3B}~\citep{yang2025qwen3}, covering different model scales and MoE architectures. This wide set of models ensures a comprehensive evaluation across scales and model families.

\paragraph{Baselines.}
We compare \method against five baselines covering the dominant families of PEFT-for-MoE: PERFT-E~\citep{liu2024perft} (per-expert LoRA following the base routing), MoELoRA~\citep{luo2024moelora} (per-layer LoRA expert pool with an independent learned router),
MoLA~\citep{gao2025mola} (heuristic expert allocation for MoE-style LoRA), DAS-LoRA~\citep{tang2026exploring} (DAS-guided static selection of a subset of base experts for
adaptation), and FFT (full fine-tuning). All methods are applied only to the MoE modules while keeping all other modules frozen.

\paragraph{Datasets.}
We organize datasets by domain. \emph{Math:} fine-tune on
MetaMathQA-R1~\citep{yu2024metamath}; evaluate on
GSM8K~\citep{cobbe2021training} and MATH-500~\citep{lightman2023lets}.
\emph{Code:} fine-tune on MagicCoder-OSS~\citep{wei2023magicoder}
($75$K samples); evaluate on MBPP~\citep{austin2021program} and
HumanEval~\citep{chen2021evaluating}. \emph{General retention:} evaluation on five general-domain benchmarks (MMLU\citep{mmlu2021}, WinoGrande\citep{sakaguchi2021winogrande}, ARC-Challenge\citep{arc2018}, StrategyQA\citep{geva2021strategyqa}, and CommonsenseQA\citep{talmor2019commonsenseqa}) using the code fine-tuned checkpoints to measure general capability retention after domain-specific adaptation.

\paragraph{Fair comparison.}
We keep all essential evaluation criteria identical for all evaluated baselines: sharing the same training and testing recipe. For all evaluated methods except FFT, the trainable parameter budgets vary by at most ${\sim}10\%$. (per-method counts and configurations are listed in Appendix~\ref{app:hparams}, and training-time and memory comparisons with PEFT baselines are reported in Appendix~\ref{app:training_efficiency}).

\subsection{Main Experimental Results}

Tab.~\ref{tab:main} reports in-domain performance
(Math and Code) together with out-of-domain general
retention across three MoE backbones.

Across all three backbones, \method achieves the
highest in-domain average among PEFT methods while
maintaining strong general capability retention after
domain-specific fine-tuning. The gains remain
consistent across models with different scales and
expert granularities, ranging from OLMoE-1B-7B to
Qwen3-30B-A3B. In particular, \method improves over
prior PEFT-on-MoE methods on both math and code
benchmarks simultaneously, whereas several baselines
show a stronger trade-off between in-domain adaptation
and out-of-domain retention. 

\begin{table*}[t]
\centering
\footnotesize
\setlength{\tabcolsep}{3pt}

\resizebox{\textwidth}{!}{%
\begin{tabular}{l cccc|c|ccccc|c|}
\toprule
& \multicolumn{5}{c}{\textbf{In-domain (Math + Code)}}
& \multicolumn{6}{c}{\textbf{General Retention}} \\
\cmidrule(lr){2-6} \cmidrule(lr){7-12}
\textbf{Method} & GSM8K & MATH & MBPP & HE & \textbf{Avg} & MMLU & WG & ARC-C & SQA & CSQA & \textbf{Avg} \\
\midrule
\multicolumn{12}{l}{\textit{OLMoE-1B-7B\cite{olmoe2024}}} \\
Base                                    & 1.06  & 0.2   & 23.01 & 14.23 & 9.63  & 45.08 & 51.22 & 52.84 & 56.74 & 45.54 & 50.28 \\
FFT                                     & 46.70 & 17.2  & 28.79 & 21.75 & 28.61 & 45.01 & 50.67 & 52.05 & 53.42 & 48.48 & 49.93 \\
per-expert LoRA\cite{liu2024perft}      & 29.11 & 7.6   & 28.79 & 14.43 & 19.98 & 48.44 & 50.99 & 56.83 & 52.11 & 47.58 & 51.19 \\
MoLA\cite{gao2025mola}                     & 26.38 & 7.8   & 29.18 & 15.24 & 19.65 & \textbf{48.58} & 51.22 & 57.42 & 50.80 & 46.68 & 50.94 \\
MoELoRA\cite{luo2024moelora}            & 29.91 & 5.6   & 29.57 & 15.85 & 20.23 & 47.71 & 51.46 & 56.48 & \textbf{53.86} & 45.21 & 50.94 \\
DAS-LoRA\cite{tang2026exploring}        & 34.87 & 8.2 & 29.96 & 16.46 & 22.37 & 47.82 & \textbf{51.54} & 57.68 & 52.84 & 45.37 & 51.05 \\
\textbf{\method}                        & \textbf{36.32}& \textbf{11.2} & \textbf{31.91} & \textbf{16.87} & \textbf{24.07}& 48.50 & 51.46 & \textbf{58.19} & 53.28 & 46.60 & \textbf{51.61} \\
\midrule
\multicolumn{12}{l}{\textit{DeepSeek-V2-Lite\cite{deepseekv2}}} \\
Base                                    & 3.71  & 0.0   & 26.46 & 18.90 & 12.27 & 46.09 & 49.96 & 55.03 & 49.34 & 42.59 & 48.60 \\
FFT                                     & 65.05 & 26.0  & 47.86 & 40.45 & 44.84 & 51.55 & 52.17 & 64.42 & 49.49 & 50.12 & 53.55 \\
per-expert LoRA\cite{liu2024perft}      & 49.28 & 14.4  & 47.86 & 32.32 & 35.97 & 51.59 & 51.30 & 64.85 & 49.20 & 49.80 & 53.35 \\
MoLA\cite{gao2025mola}                     & 46.25 & 13.8  & 49.03 & \textbf{34.15} & 35.81 & 51.60 & 50.12 & 64.33 & 49.20 & \textbf{52.42} & 53.53 \\
MoELoRA\cite{luo2024moelora}            & 48.52 & 14.6  & 49.81 & 33.13 & 36.52 & 51.22 & 50.91 & 64.42 & \textbf{51.53} & 49.96 & 53.61 \\
DAS-LoRA\cite{tang2026exploring}        & 46.55 & 14.6  & 49.03 & 33.54 & 35.93 & 49.64 & 51.54 & 61.77 & 51.09 & 49.80 & 52.77 \\
\textbf{\method}                        & \textbf{52.24}& \textbf{18.4}& \textbf{52.14} & 33.54 & \textbf{39.08}& \textbf{51.72} & \textbf{51.85} & \textbf{65.53} & 50.22 & 52.01 & \textbf{54.27} \\
\midrule

\multicolumn{12}{l}{\textit{Qwen3-30B-A3B\cite{yang2025qwen3}}} \\
Base                                    &  73.69 & 30.6 & 69.26  & 63.41 & 59.24 & 77.77 & 71.82 & 92.83 & 72.49 & 85.34 & 80.05 \\
FFT                                     & 95.30 & 74.40 & 72.76 & 82.52 & 81.23 & 74.46 & 71.35 & 92.75 & 61.86 & 81.98 & 76.48 \\
per-expert LoRA\cite{liu2024perft}      & 94.47 & 71.60 & 72.37 & 82.11 & 80.14 & 75.05 & 72.38 & 93.34 & 70.16 & 84.03 & 78.99 \\
MoLA\cite{gao2025mola}                     & 95.07 & 72.60 & 71.98 & 82.11 & 80.44 & 73.79 & \textbf{72.77} & 92.24 & 71.03 & 82.72 & 78.51 \\
MoELoRA\cite{luo2024moelora}            & 94.84 & 72.00 & 73.54 & \textbf{82.52} & 80.73 & 68.17 & 71.03 & 86.01 & 72.05 & 75.43 & 74.54 \\
DAS-LoRA\cite{tang2026exploring}        & 94.69 & 73.20 & 73.93 & 81.71 & 80.88 & 69.18 & 71.11 & 87.80 & \textbf{73.65} & 84.03 & 77.15 \\
\textbf{\method}                        & \textbf{95.22} & \textbf{74.40} & \textbf{74.32} & 82.32 & \textbf{81.57} & \textbf{76.21} & 71.51 & \textbf{93.60} & 70.01 & \textbf{85.83} & \textbf{79.43} \\

\bottomrule
\end{tabular}
}

\caption{
Main results across three MoE backbones on
in-domain tasks (Math, Code) and out-of-domain
general retention. Avg denotes the combined in-domain average
over GSM8K, MATH, MBPP, HumanEval. Bold indicates the best PEFT
result excluding FFT.
}
\label{tab:main}
\end{table*}

\subsection{Multimodal Experiments}
\label{sec:mm}

To evaluate whether \method remains effective in multimodal settings, we further conduct experiments on a vision-language MoE model in the medical imaging domain.
\paragraph{Setup.}
We evaluate \method and two representative baselines on \textbf{Qwen3.5-35B-A3B}~\citep{qwen3.5}.
For in-domain
medical VQA (Vision Question Answering) we fine-tune on a mixture of LLaVA-Med~\citep{li2024llavamed} and the
train splits of three medical VQA datasets, and evaluate on the test splits of VQA-RAD~\citep{lau2018dataset},
SLAKE~\citep{liu2021slake}, and PathVQA~\citep{he2020pathvqa}. For
\emph{general VL retention}, we evaluated the checkpoints trained on medical VQA datasets on
MMBench~\citep{liu2024mmbench}, MME~\citep{fu2023mme}, and
RealWorldQA~\citep{realworldqa}.

\paragraph{Results.}
Tab.~\ref{tab:medical} reports both axes. \method achieves the best accuracy on every in-domain benchmark and the best positive transfer on every general benchmark, demonstrating the effectiveness of \method on multimodal settings.

\begin{table*}[t]
\centering
\small
\setlength{\tabcolsep}{3.5pt}
\begin{tabular}{l ccc |c| c ccc |c|}
\toprule
& \multicolumn{4}{c}{\textbf{Medical VQA}} & & \multicolumn{4}{c}{\textbf{General Retention}} \\
\cmidrule(lr){2-5} \cmidrule(lr){7-10}
\textbf{Method} & \textbf{VQA-RAD} & \textbf{SLAKE} & \textbf{PathVQA} & \textbf{Avg} & & \textbf{MMBench} & \textbf{MME} & \textbf{RealWorldQA} & \textbf{Avg} \\
\midrule
\multicolumn{10}{l}{\textit{Qwen3.5-35B-A3B\cite{qwen3.5}}} \\
Base model                              & 69.32 & 72.28 & 50.47 & 64.02 & & 90.14 & 91.03 & 46.41 & 75.86 \\
MoELoRA\cite{luo2024moelora}            & 68.20 & 87.30 & 64.82 & 73.44 & & 91.61 & 92.04 & 46.14 & 76.60 \\
DAS-LoRA\cite{tang2026exploring}        & 68.60 & 87.40 & 64.60 & 73.53 & & 91.61 & 92.38 & 46.54 & 76.84 \\
\textbf{\method}                        & \textbf{69.25} & \textbf{87.68} & \textbf{65.67} & \textbf{74.20} & & \textbf{92.19} & \textbf{92.42} & \textbf{47.19} & \textbf{77.27} \\
\bottomrule
\end{tabular}
\caption{Medical VQA accuracy and general VL retention on
Qwen3.5-35B-A3B.}
\label{tab:medical}
\end{table*}

\subsection{Experiments on Capacity Scaling}

We study scaling behavior under different trainable
parameter budgets ($0.12\%$, $0.24\%$, and
$0.48\%$) on OLMoE math
(Tab.~\ref{tab:capacity}). As the trainable budget
increases, \method consistently improves on
GSM8K, rising from $0.3707$ at $0.12\%$ to
$0.3980$ at $0.48\%$. In contrast, MoELoRA shows
limited gains under larger budgets, particularly on
GSM8K, where performance remains around $0.30$
across all settings.

The results show that \method maintains stronger
scaling behavior than MoELoRA under increased
adaptation capacity.


\begin{table}[t]
\centering
\small
\resizebox{\columnwidth}{!}{
\begin{tabular}{lccc}
\toprule
\textbf{Method} & \textbf{Params} & \textbf{GSM8K} & \textbf{MATH-500} \\
\midrule
MoELoRA & 0.12\% & 29.91 & 5.6 \\
MoELoRA & 0.24\% & 30.10 & 8.4 \\
MoELoRA & 0.48\% & 30.02 & 9.4 \\
\midrule
\method & 0.12\% & 37.07 & 8 \\
\method & 0.24\% & 39.58 & 9.2 \\
\method & 0.48\% & 39.8  & 10.6 \\
\bottomrule
\end{tabular}}
\caption{Capacity scaling on OLMoE math. \method scales more effectively than MoELoRA.}
\label{tab:capacity}
\vspace{-3mm}
\end{table}

\section{Analysis and Discussion}
\label{sec:ablation}

In this section, we first implement ablation studies on \method to evaluate the effectiveness of its components. We then conduct in-depth analyses on the working mechanisms of the RCP and the global expert pool, uncovering the underlying advantages of our designs.

\subsection{Component Ablation}
\label{sec:component_ablation}

We isolate \method's two architectural choices, routing-conditioned projection (RCP) and the shared global pool,through a matched-budget ablation chain on DeepSeek-V2-Lite (math). Starting from a
MoELoRA-style baseline, we first replace
hidden-state routing with RCP while keeping the
per-layer pool fixed, and then introduce the
globally shared pool on top of RCP.

\begin{table*}[t]
\centering
\small
\begin{tabular}{lccccc}
\toprule
\textbf{Variant}
& \textbf{Pool}
& \textbf{Routing}
& \textbf{GSM8K}
& \textbf{MATH-500}
& \textbf{Avg} \\
\midrule

Baseline
& per-layer
& hidden-state
& 48.52
& 14.6
& 31.56 \\

+ RCP
& per-layer
& RCP
& 49.05
& 15.4
& 32.23 \\

+ Global Pool (\method)
& global
& RCP
& \textbf{52.24}& \textbf{18.4}& \textbf{35.32}\\

\midrule

Global Only
& global
& hidden-state
& 48.07
& 16.2
& 32.14 \\

Router-only (per-layer)
& per-layer
& $z$ only
& 45.49
& \textbf{16.8}
& 31.15 \\

Router-only (global)
& global
& $z$ only
& 46.17
& 15.8
& 30.99 \\

\bottomrule
\end{tabular}

\caption{
Component ablation on DeepSeek-V2-Lite math.
}
\label{tab:component}
\end{table*}

Tab.~\ref{tab:component} shows the main ablation
chain (top three rows) together with additional
reference variants (bottom). Replacing
hidden-state routing with RCP under a fixed
per-layer pool improves Avg from $31.56$ to
$32.23$, showing that incorporating pretrained
router structure improves LoRA expert selection
even without cross-layer sharing. Introducing the
globally shared pool on top of RCP further
improves Avg to $35.32$, indicating that global sharing and routing-conditioned projection are complementary.

We further ablate the $h_c$ (Auxiliary channel) by routing only from pretrained router logits
$z \in \mathbb{R}^{N_E}$. Both router-only
variants underperform the full RCP design,
indicating that pretrained routing structure
alone is insufficient. Effective LoRA routing
requires combining inherited router priors with input-dependent adaptation signals from the hidden state.

\subsection{Analysis on RCP}
\label{sec:analysis_rcp}

\begin{figure}[t]
\centering
\includegraphics[width=\linewidth]{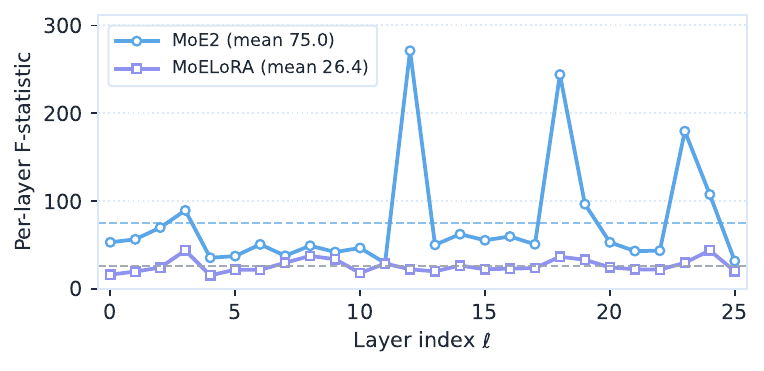}
\caption{Per-layer routing-alignment F-statistic on DeepSeek-V2-Lite:
\method (red) versus the independent-routing baseline (gray). }
\label{fig:fstat_per_layer}
\vspace{-3mm}
\end{figure}

To verify that RCP's per-layer projection $W_\ell$ actually inherits
information from the base MoE router rather than learning an
independent routing function, we measure the F-statistic of structural
alignment between LoRA routing and base routing, with details provided
in Appendix~\ref{app:fstat}. A larger F-statistic indicates
that LoRA expert selection varies more systematically with the base
router's expert assignments, and therefore better reflects the routing
structure  learned by the pretrained MoE model.

On DeepSeek-V2-Lite, \method achieves $F = 75.0$, compared with
$F = 26.4$ for MoELoRA, yielding a $2.8\times$ margin.
Fig.~\ref{fig:fstat_per_layer} further breaks down this comparison
across layers. \method obtains a higher F-statistic on all 26 layers. This
layer-wise consistency suggests that the alignment does not come from
a few isolated layers, but is a systematic effect of conditioning LoRA
routing on pretrained router logits.

These results show that RCP produces LoRA routing patterns that are
substantially more aligned with the base MoE routing structure than
MoELoRA's independent router. This provides direct evidence that the
pretrained router logits are effectively reflected in adapter selection,
supporting the intended role of RCP as a router-conditioned projection
rather than an independently learned LoRA router.


\subsection{Analysis on Global Pool}
\label{sec:analysis_global}

The shared pool architecture imposes no explicit
depth constraint on expert usage. We analyze the
resulting organization of the trained pool on
DeepSeek-V2-Lite, focusing on its layer-aware depth organization
(\S\ref{sec:depth_organized}), non-uniform capacity
allocation (\S\ref{sec:capacity_nonuniform}), and
the cross-layer representational overlap that may
support such sharing (\S\ref{sec:cka_substrate}).

For the trained-model probes, we run forward
  passes of the trained \method on 300 GSM8K
  test prompts and collect the set $T$ of all
  token positions. For each layer $\ell$ and
  LoRA expert $e$, we record the top-$K_L$
  selection frequency 
  \[
  f_{\ell,e} \;=\; \frac{1}{|T|}
  \sum_{t \in T} \mathbb{1}\!\bigl[
  e \in \mathcal{S}_{h_t}^{\mathrm{LoRA}}\bigr],
  \]
  where $\mathcal{S}_{h_t}^{\mathrm{LoRA}}$ is
  the activated LoRA set for token $t$ at
  layer $\ell$ (\S\ref{sec:wl}). We use the
  discrete top-$K_L$ frequency rather than the
  soft routing weight to align with the
  experts that the model actually engages at
  inference.

\subsubsection{Global Pool Learns Layer-aware Expert Assignment}
\label{sec:depth_organized}

A potential concern of a shared global LoRA pool is that
experts may be used in an unstructured manner across layers,
without reflecting layer-specific adaptation behavior.
We therefore analyze whether the learned routing exhibits
depth-dependent organization. For each expert $e$, we compute its preferred depth
\[
\bar\ell_e =
\frac{\sum_\ell \ell \cdot f_{\ell,e}}
     {\sum_\ell f_{\ell,e}},
\]
where $f_{\ell,e}$ denotes the activation frequency of
expert $e$ at layer $\ell$.
Experts are then sorted by $\bar\ell_e$.

Fig.~\ref{fig:overview}(III) visualizes the activation matrix
$f_{\ell,e}$ defined above, with MoE layers on the x-axis,
LoRA experts sorted by preferred depth $\bar\ell_e$ on the
y-axis, and color intensity denoting activation frequency.
The resulting diagonal structure, quantified by a Spearman
correlation of $\rho = 0.92$ between $\bar\ell_e$ and each
expert's most frequently activated layer, indicates that
expert usage is depth-structured rather than layer-agnostic.

Notably, the resulting structure is localized rather than
strictly layer-exclusive.
Individual experts typically concentrate their activation
mass within a narrow band of neighboring layers, instead of
being activated by only a single layer.
As a result, the global pool simultaneously preserves
layer-dependent specialization and enables cross-layer
expert reuse.

Appendix~\ref{app:expert_affinity_robust} shows that this
depth-organized structure remains consistent across
different pool sizes
$N_L \in \{128,256,512\}$
and across both math and code probes
(Figs.~\ref{fig:expert_affinity_pool_size},
\ref{fig:expert_affinity_math_vs_code}).

\subsubsection{Per-layer Expert Capacity Is Non-uniform}
\label{sec:capacity_nonuniform}

\begin{figure}[t]
\centering
\includegraphics[width=\linewidth]{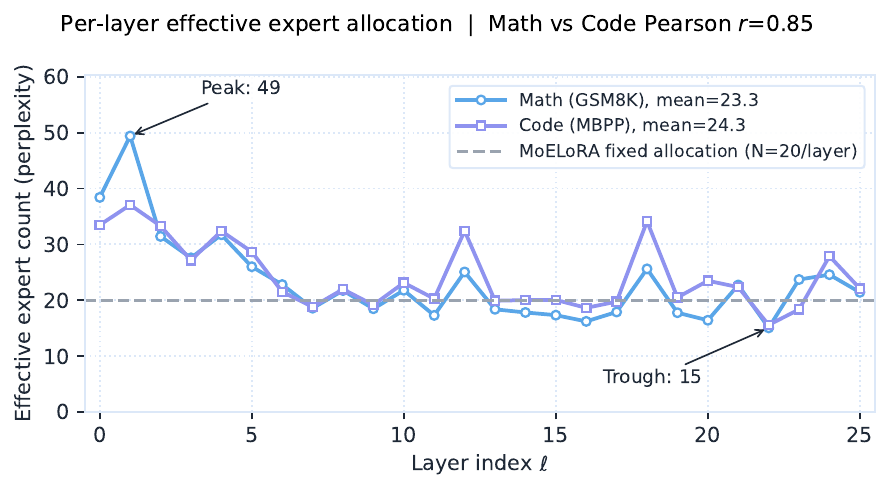}
\caption{Per-layer effective expert count on math (blue) and code
(orange) probes. Dashed line: the fixed allocation used by MoELoRA.}
\label{fig:layer_allocation}
\vspace{-3mm}
\end{figure}

For each layer $\ell$, we compute the effective expert count as the
perplexity of its usage distribution, $\exp(H(f_\ell))$.
Fig.~\ref{fig:layer_allocation} shows that the learned allocation is
highly non-uniform: the effective expert count ranges from $\approx 15$
in the deepest layers to $\approx 49$ in the shallowest layers, a
$3.3\times$ variation.

This result suggests that adaptation demand varies substantially across
depth, which is difficult to capture with fixed per-layer allocation.
MoELoRA assigns the same number of experts to each layer
($N{=}20$; dashed line), while MoLA requires manually specifying a
layer-wise schedule such as bottom-heavy, top-heavy, or hourglass. In
contrast, \method exposes no explicit architectural knob for per-layer
capacity, yet learns a non-uniform allocation automatically.

\subsubsection{Cross-layer Representational Overlap Supports Pool Sharing}
\label{sec:cka_substrate}

We further examine why sharing LoRA experts across layers is feasible.
Specifically, we measure linear CKA(Appendix~\ref{app:cka}) between MoE block outputs of every
layer pair on the frozen DeepSeek-V2-Lite base model over $200$ GSM8K
prompts.

\begin{figure}[t]
\centering
\includegraphics[width=\linewidth]{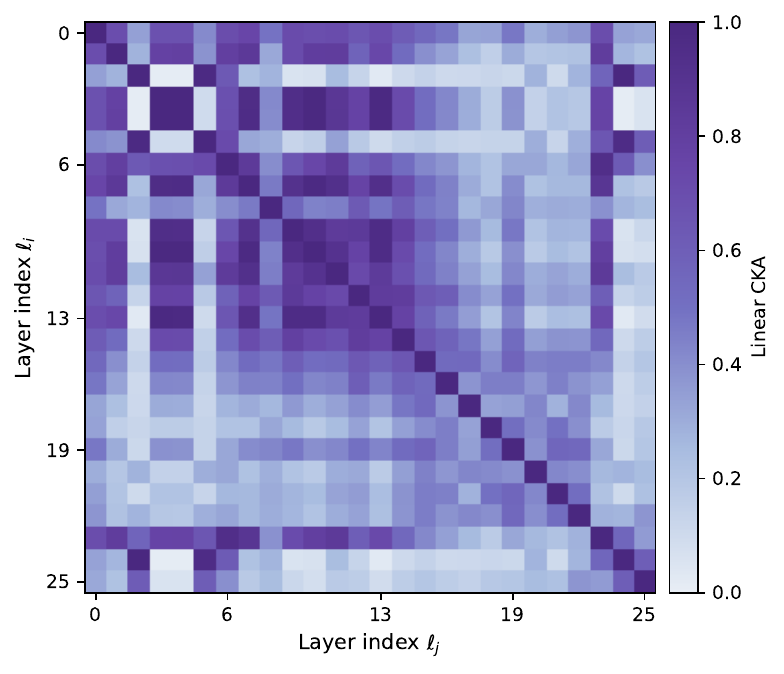}
\caption{Linear CKA between MoE block outputs across the
$26$ layers of DeepSeek-V2-Lite (base model, no LoRA).}
\label{fig:cka}
\vspace{-3mm}
\end{figure}

Fig.~\ref{fig:cka} shows clear cross-layer similarity: off-diagonal CKA
averages $0.43$, with adjacent layers averaging $0.56$ compared with
$0.35$ for distant pairs ($|i{-}j|{\geq}10$). This depth-local
representational overlap provides a plausible basis for sharing LoRA
experts across nearby or related layers, consistent with the soft
depth-affinity pattern observed above.

\section{Conclusion}

We introduced \method, an MoE-style PEFT method that
deeply binds the MoE mechanisms of pretrained MoE models
and MoE-style low-rank adaptation through
routing-conditioned projection (RCP) and a globally
shared LoRA expert pool. Across four MoE backbones
spanning different scales and expert granularities,
\method achieves strong downstream performance
while preserving general capabilities under matched
training budgets. Further analysis shows that \method learns structured cross-layer adaptation while preserving alignment with pretrained MoE routing, confirming the effectiveness of its model-level design. Together, these results show that \method is an
effective approach to PEFT for MoE models.

\section*{Limitations}

\paragraph{Adapter merging is not straightforward.}
Like other routing-based PEFT methods such as
MoELoRA, \method relies on dynamic token-wise
expert selection rather than a fixed low-rank
update. As a result, the learned adaptation cannot
be directly merged into the base weights in the
same manner as standard LoRA.

Developing mergeable approximations for dynamic
routing-based PEFT architectures remains an
important direction for future work.

\paragraph{Scaling the global pool increases projection parameter cost.}

The globally shared expert pool improves adaptation
flexibility, but scaling the pool size also increases
the parameter cost of the routing/projection
interface, particularly the layer-specific projection
matrices used for expert selection. A naive design
that routes directly from hidden states would cause
these projection parameters to grow rapidly with
both hidden dimension and pool size.

\method mitigates this issue by performing routing
in the pretrained router space rather than the full
hidden-state space, substantially reducing projection
parameter growth while preserving routing
expressiveness. Nevertheless, scaling to extremely
large shared pools may still require more
parameter-efficient routing mechanisms.

\bibliography{references}

\appendix

\section{Reproducibility Details}
\label{app:repro}

\subsection{Evaluation Protocol}
\label{app:eval_protocol}

This appendix specifies the evaluation
protocol used in all main-text experiments.
The protocol is shared verbatim across
\method and all baselines; only the trained
adapter weights differ.

\paragraph{Inference configuration.}
The decoding configuration is shared across
all methods within each benchmark; only the
trained adapter weights differ. All benchmarks
are evaluated in a strictly zero-shot setting
without in-context examples. GSM8K, MATH-500,
and MBPP use greedy decoding
(\texttt{do\_sample=False},
\texttt{temperature=0}), while HumanEval uses
stochastic decoding with
\texttt{temperature=0.2} and
\texttt{top\_p=0.95}. We use
\texttt{max\_new\_tokens=1024} for all
benchmarks. MBPP is evaluated with pass@1,
and for HumanEval we generate $n{=}3$ samples
per problem and report the average pass@1
across runs to reduce variance on the small
164-task test set.

\paragraph{Prompt templates.}
We use a single fixed prompt template per
benchmark, applied identically to every
method:
\begin{itemize}
  \item \textbf{GSM8K}:
    \texttt{"Solve the following math
    problem step by step. End your answer
    with `\#\#\#\# <number>'.\textbackslash
    n\textbackslash nQuestion:
    \{question\}\textbackslash nAnswer:"}
  \item \textbf{MATH-500}:
    \texttt{"Solve the following math
    problem step by step. Put your final
    answer in \textbackslash boxed\{...\}.
    \textbackslash n\textbackslash nProblem:
    \{problem\}\textbackslash nSolution:"}
  \item \textbf{MBPP / HumanEval}: standard
    function-completion prompts (problem
    description plus public test signature);
    the model is asked to produce the
    function body, which is then executed
    against the held-out unit tests.
\end{itemize}

\paragraph{Answer extraction.}
Answer extraction is identical across all
methods. For GSM8K, we extract the final
numeric answer using a regex match against
\texttt{\#\#\#\#\textbackslash s*(-?[0-9.,]+)}
and evaluate after numeric normalization.
For MATH-500, we extract the contents of the
first \texttt{\textbackslash boxed\{\ldots\}}
expression (including nested braces) and
evaluate after whitespace and LaTeX-symbol
normalization. For MBPP and HumanEval, the
generated function body is parsed from the
completion and executed against the official
unit tests, with pass@1 reported following
the standard benchmark protocols.

\subsection{Hyperparameters}
\label{app:hparams}

\paragraph{\method.}
Table~\ref{tab:hparams_ours} summarizes the key
hyperparameters used for \method across different
backbones. Here, $N_L$ denotes the size of the
shared LoRA expert pool, rank the per-expert LoRA
rank, top-$K$ the number of activated LoRA experts
per token, and $d_b$ the hidden-state bottleneck
dimension used in routing-conditioned projection.

\begin{table*}[t]
\centering
\small
\begin{tabular}{lccccc}
\toprule
\textbf{Backbone} & $N_L$ & rank & top-$K$ & $d_b$ & \#params \\
\midrule
OLMoE-1B-7B      & 128 & 16 & 4 & 16 & 9.1M \\
DeepSeek-V2-Lite & 256 & 16  & 6 & 64 & 21M \\
Qwen3-30B-A3B    & 256 & 32 & 6 & 32 & 39M \\
\bottomrule
\end{tabular}
\caption{\method hyperparameters across backbones.
On Qwen3-30B-A3B the pool is further partitioned
into $G{=}2$ groups (hierarchical routing); the
other two backbones use a single flat pool
($G{=}1$).}
\label{tab:hparams_ours}
\end{table*}

\paragraph{MoELoRA~\citep{luo2024moelora}.}
We use a per-layer LoRA expert pool with dynamic
routing from hidden states. The configuration uses
16 experts (rank 8, top-$2$) on OLMoE-1B-7B,
32 experts (rank 6, top-$4$) on
DeepSeek-V2-Lite, and 32 experts (rank 6,
top-$4$) on Qwen3-30B-A3B, corresponding to
approximately $8.9$M, $21$M, and $41$M trainable parameters.

\paragraph{MoLA~\citep{gao2025mola}.}
MoLA uses a per-layer LoRA expert pool with a
layer-wise increasing allocation, motivated by
the observation that deeper transformer layers
benefit from more LoRA experts. LoRA experts are
routed independently from hidden states
(top-$2$ per token). Following the original
paper's allocation scheme, we use ranks $8$,
$22$, and $40$ on OLMoE-1B-7B,
DeepSeek-V2-Lite, and Qwen3-30B-A3B
respectively, with per-layer expert counts
ranging from $8$ to $24$ (OLMoE, 4 stages),
$4$ to $16$ (DeepSeek, 26 layers), and $2$ to
$8$ (Qwen3, 4 stages). On DeepSeek-V2-Lite we
scale MoLA's rank from the original $16$ to
$22$ to match the parameter budget of other
baselines. This yields approximately $8.9$M,
$21.4$M, and $39$M trainable parameters.

\paragraph{DAS-LoRA~\citep{tang2026exploring}.}
For DAS-LoRA, we retain the original
CDAS-based expert selection procedure but
restrict training to the selected LoRA
modules only, without additionally tuning
router or dense backbone parameters, in
order to maintain comparable PEFT settings
across methods. Following the original paper, we pre-select a subset of base experts using Domain Advantage Score and attach LoRA modules only to the selected experts. We use rank $32$ on OLMoE-1B-7B, $52$ on DeepSeek-V2-Lite, and $48$ on Qwen3-30B-A3B, yielding approximately $8.4$M, $22$M, and $38$M trainable parameters.

\paragraph{Training.}
Within each backbone, all PEFT methods share the
same optimizer (AdamW), cosine learning-rate
schedule with warmup ratio $0.1$, batch size,
and number of epochs. Method-specific
hyperparameters are tuned on the validation split
under comparable trainable parameter budgets.

\subsection{Training Compute}

All experiments are conducted on a single
8$\times$NVIDIA H800 80GB node using bfloat16
training. DeepSeek-V2-Lite and OLMoE use standard
DDP, while Qwen3-30B-A3B uses DeepSpeed ZeRO-3.
Typical wall-clock training time per run is
approximately 35 minutes for OLMoE, 1.5 hours for
DeepSeek-V2-Lite, and 2.5--3 hours for
Qwen3-30B-A3B. The total compute budget for all
experiments is approximately 200 GPU-hours.

\paragraph{Reproducibility note.}
Each adapter is trained once per (method,
backbone) configuration. We report the
per-benchmark scores from that run directly,
following the evaluation protocol in
Appendix~\ref{app:eval_protocol}: a single
greedy sample for GSM8K, MATH-500 and MBPP,
and $n{=}3$ stochastic samples (temperature
$0.2$) averaged for HumanEval. Multi-run
training-time statistics are not reported
due to compute cost.

\subsection{Training-time and memory comparability with PEFT baselines}
\label{app:training_efficiency}

Table~\ref{tab:training_efficiency} compares
training time and peak training memory on
DeepSeek-V2-Lite using 80k samples extracted
from MetaMathQA for one training epoch. Training time is measured on $8\times$ H800 80GB GPUs with DDP, while peak memory is measured on a single H800 without ZeRO using batch size $1$ and sequence length $2048$. All PEFT baselines are matched to roughly $22$M trainable parameters; per-expert
LoRA is included for reference.

  \begin{table*}[t]
  \centering\small
  \setlength{\tabcolsep}{4pt}
  \begin{tabular}{lcccc}
  \toprule
  Method & \makecell{Trainable\\Params} & \makecell{Train\\Time (min)} & \makecell{Peak Train\\Mem (GB)} \\
  \midrule
  \textbf{\method}             & 21.7M & 16.6 & 44.65 \\
  MoELoRA~\citep{luo2024moelora}             & 22.2M & 19.8 & 43.6 \\
  MoLA~\citep{gao2025mola}                   & 21.4M & 17.2 & 42.7  \\
  DAS-LoRA~\citep{tang2026exploring}         & 22.2M & 17.8 & 41.8  \\
  per-expert LoRA~\citep{liu2024perft}$^{\dagger}$    & 27.3M & 23.2 & 44.3 \\
  \midrule
  FFT (DeepSeek-V2-Lite reference)$^{\ddagger}$              & 16B   & 143  & 120 \\
  \bottomrule
  \end{tabular}
  \caption{Training-time efficiency on DeepSeek-V2-Lite math.
  $^{\dagger}$Per-expert LoRA cannot be configured to a 22M
  budget without collapsing to rank-1; we report its native
  27M configuration.}
  \label{tab:training_efficiency}
  \end{table*}

All PEFT methods exhibit broadly comparable
training-time and memory cost, with training
time falling within a narrow $17$--$20$ minute
range and peak memory between $41.8$--$46.4$
GB. \method lies slightly above the lightest
baselines in both metrics, largely because it
activates more LoRA experts per token.
Compared with per-expert LoRA, the globally
shared pool also avoids the larger training
overhead associated with materializing
independent LoRA modules for every expert.

\section{Additional Analysis Details}
\label{app:analysis_details}

\subsection{Robustness of the Depth-Organized Expert Allocation}
\label{app:expert_affinity_robust}

To verify that the depth-organized diagonal observed in
Section~\ref{sec:depth_organized} is not specific to a
particular configuration, we reproduce the analysis across
(i) different global pool sizes and (ii) different probe
domains.

\begin{figure}[t]
\centering
\includegraphics[width=\linewidth]{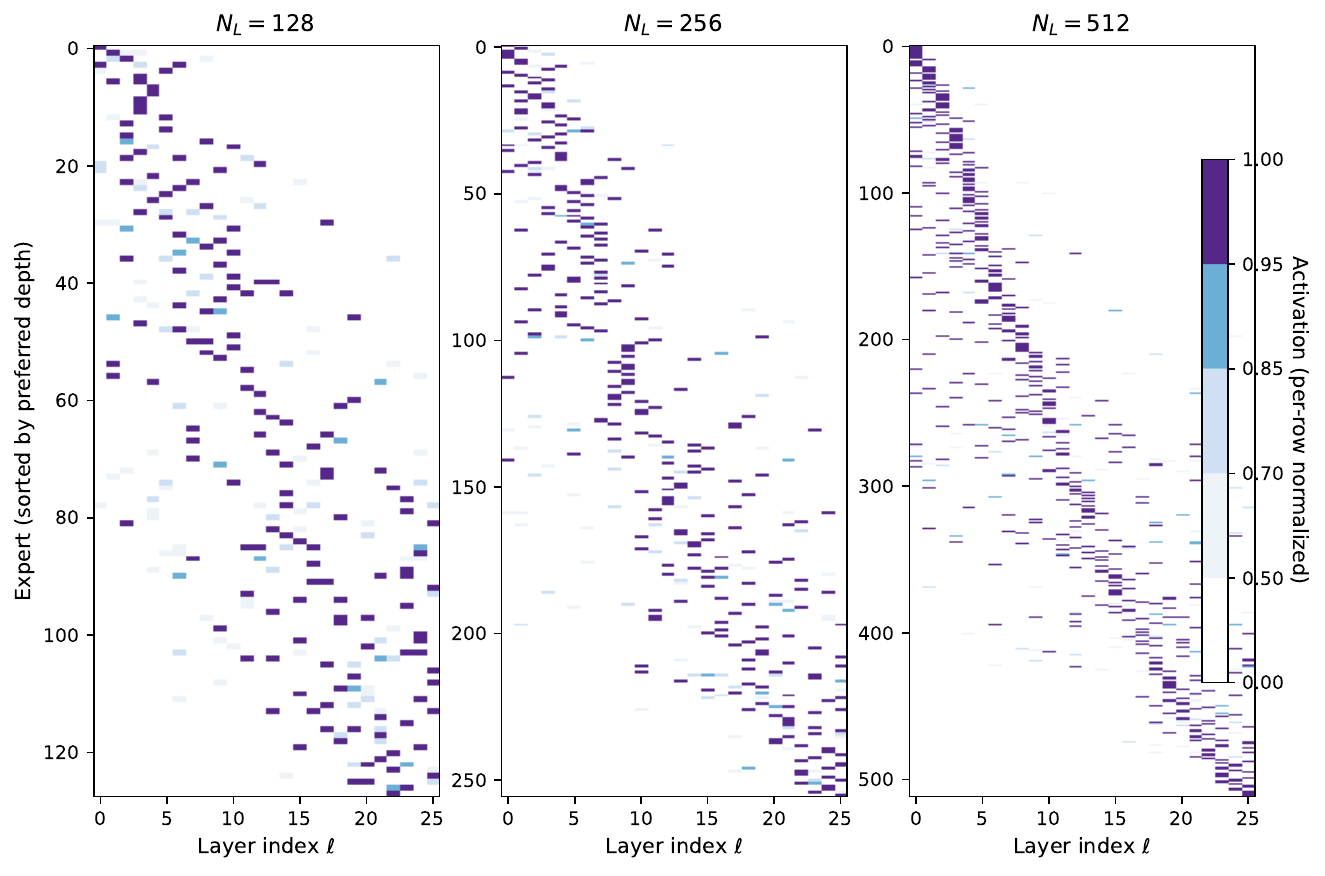}
\caption{Expert-affinity matrices on GSM8K for
\method with different global pool sizes
$N_L \in \{128,256,512\}$ on DeepSeek-V2-Lite.
Rows are experts sorted by preferred depth; columns are
MoE layers. A consistent depth-organized diagonal emerges
across all pool sizes.}
\label{fig:expert_affinity_pool_size}
\end{figure}

\paragraph{Across pool sizes.}
Fig.~\ref{fig:expert_affinity_pool_size} shows the
expert-affinity matrices on GSM8K for
$N_L \in \{128, 256, 512\}$.
The same depth-organized diagonal appears in all three
settings, indicating that experts consistently specialize
to contiguous depth regions.
As $N_L$ increases, the diagonal becomes thinner because
the 26-layer depth axis is partitioned among more experts;
smaller pools produce correspondingly broader depth bands.

\begin{figure}[t]
\centering
\includegraphics[width=\linewidth]{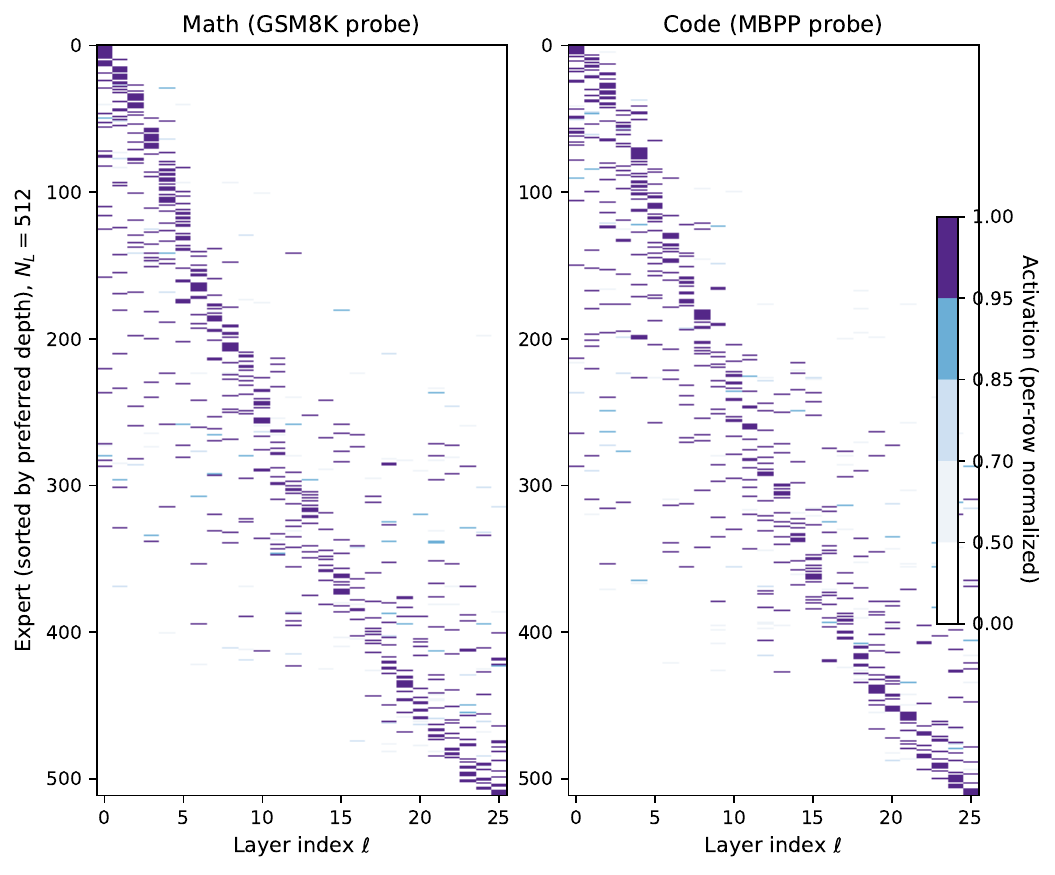}
\caption{Expert-affinity matrices for the same trained
adapter evaluated on GSM8K (left) and MBPP (right).
Each panel is independently sorted by preferred depth.
The depth-organized structure is preserved across domains.}
\label{fig:expert_affinity_math_vs_code}
\end{figure}

\paragraph{Across domains.}
Fig.~\ref{fig:expert_affinity_math_vs_code}
compares GSM8K and MBPP using the same trained adapter
($N_L = 512$).
Both domains exhibit the same depth-organized structure,
despite the code probe appearing slightly noisier due to
the smaller MBPP evaluation set.

Across both pool sizes and probe domains, the same
depth-organized allocation consistently emerges,
suggesting that it is an intrinsic property of the shared
global pool with routing-conditioned projection rather
than an artifact of a particular setting.

\subsection{F-statistic for routing alignment}
\label{app:fstat}

The routing-alignment F-statistic reported in Section~\ref{sec:analysis_rcp}
is a one-way ANOVA statistic that measures how much of the variance in
LoRA routing can be explained by the base MoE router's expert
assignment---higher values indicate stronger structural inheritance of
the pretrained routing prior.

Formally, let $T$ denote the token set, $\mathbf{p}_L(t) \in
\mathbb{R}^{N_L}$ the LoRA routing distribution for token $t$, and
$e_b(t) = \arg\max_e\, \mathbf{p}_B(t)$ the base router's top-1 expert
for $t$. We partition $T$ into $K$ groups indexed by the base expert,
$G_e = \{t \in T : e_b(t) = e\}$, with sizes $n_e = |G_e|$ (groups with
$n_e < 5$ are dropped to avoid noisy estimates). Let
$\boldsymbol{\mu}_e = \tfrac{1}{n_e}\sum_{t \in G_e} \mathbf{p}_L(t)$
denote the group-mean LoRA distribution and $\bar{\boldsymbol{\mu}} =
\tfrac{1}{|T|}\sum_t \mathbf{p}_L(t)$ the grand mean. The between- and
within-group sums of squares are
\begin{align}
\mathrm{SS}_{\text{between}} &= \sum_{e=1}^{K} n_e \,\|\boldsymbol{\mu}_e - \bar{\boldsymbol{\mu}}\|_2^2, \\
\mathrm{SS}_{\text{within}}  &= \sum_{e=1}^{K} \sum_{t \in G_e} \|\mathbf{p}_L(t) - \boldsymbol{\mu}_e\|_2^2,
\end{align}
and the F-statistic is the ratio of mean squares
\begin{equation}
F \;=\; \frac{\mathrm{SS}_{\text{between}} / (K - 1)}{\mathrm{SS}_{\text{within}} / (|T| - K)}.
\end{equation}
A larger $F$ indicates that LoRA routing distributions are tightly
clustered within each base-expert group and well-separated across
groups---i.e., the LoRA router behaves predictably as a function of the
base router. We treat $F$ as a structural-fit diagnostic for the RCP
projection $W_\ell$ rather than a quality metric; passive coupling
schemes (e.g., per-expert LoRA) attain trivially high $F$ by
construction.

\subsection{Linear CKA}
\label{app:cka}

Given centered layer representations
$X_i \in \mathbb{R}^{n \times d_i}$ and
$X_j \in \mathbb{R}^{n \times d_j}$,
we compute linear centered kernel alignment (CKA) as

\[
\mathrm{CKA}(X_i, X_j)
=
\frac{\|X_i^\top X_j\|_F^2}
{\|X_i^\top X_i\|_F \,
 \|X_j^\top X_j\|_F}.
\]

Here $\|\cdot\|_F$ denotes the Frobenius norm.
For each layer pair, $X_i$ and $X_j$ are constructed
from the corresponding MoE block outputs collected
over the evaluation prompts.

\end{document}